\documentclass[final]{cvpr}

\usepackage{times}
\usepackage{epsfig}
\usepackage{graphicx}
\usepackage{amsmath}
\usepackage{amssymb}
\usepackage{multirow}
\usepackage{bm}
\usepackage[noend]{algpseudocode}
\usepackage{algorithmicx,algorithm}

\usepackage{color}
\usepackage{paralist}
\usepackage{array}
\usepackage{enumitem}
\usepackage{multirow}

\usepackage[pagebackref=false,breaklinks=true,colorlinks,bookmarks=false]{hyperref}

\definecolor{darkgreen}{RGB}{30,150,30}
\definecolor{darkblue}{RGB}{0,0,127}
\definecolor{darkyellow}{RGB}{171,133,0}
\definecolor{darkred}{RGB}{180,20,20}
\definecolor{darkmagenta}{RGB}{127,0,127}
\definecolor{darkcyan}{RGB}{0,127,127}

\newif\ifdrafting 
\draftingtrue 

\ifdrafting
  \newcommand{\LG} [1] {\textcolor{darkgreen}{[LG: #1]}}
  \newcommand{\VJ} [1] {\textcolor{darkblue}{[VJ: #1]}}
  \newcommand{\LS} [1] {\textcolor{darkmagenta}{[LS: #1]}}
  \newcommand{\ds} [1] {\textcolor{red}{[DS: #1]}}
  \newcommand{\TODO} [1] {{\color{darkcyan}{\bf [TODO: #1]}}}

\else
  \newcommand{\LG} [1] {}
  \newcommand{\VJ} [1] {}
  \newcommand{\LS} [1] {}
  \newcommand{\ds} [1] {}  
  \newcommand{\TODO} [1] {}
  
\fi

\def\cvprPaperID{578} 
\def\confYear{CVPR 2021}

\begin{document}

\title{Adaptive Prototype Learning and Allocation for Few-Shot Segmentation}

\author{Gen Li\textsuperscript{1,3}, Varun Jampani\textsuperscript{2}, 
Laura Sevilla-Lara\textsuperscript{1}, Deqing Sun\textsuperscript{2},
Jonghyun Kim\textsuperscript{3}, Joongkyu Kim\textsuperscript{3}\thanks{Corresponding author}\\ \\
\textsuperscript{1}University of Edinburgh\quad
\textsuperscript{2}Google Research\quad
\textsuperscript{3}Sungkyunkwan University\\
}

\maketitle
\pagestyle{empty}
\thispagestyle{empty}

\begin{abstract}
Prototype learning is extensively used for few-shot segmentation.
Typically, a single prototype is obtained from the support feature by averaging the global object information.
However, using one prototype to represent all the information may lead to ambiguities.
In this paper, we propose two novel modules, named superpixel-guided clustering (SGC) and guided prototype allocation (GPA), for multiple prototype extraction and allocation.
Specifically, SGC is a parameter-free and training-free approach, which extracts more representative prototypes by aggregating similar feature vectors, while GPA is able to select matched prototypes to provide more accurate guidance.
By integrating the SGC and GPA together, we propose the Adaptive Superpixel-guided Network (ASGNet), which is a lightweight model and adapts to object scale and shape variation.
In addition, our network can easily generalize to k-shot segmentation with substantial improvement and no additional computational cost.
In particular, our evaluations on COCO demonstrate that ASGNet surpasses the state-of-the-art method by 5\% in 5-shot segmentation.\footnote{Code is available at \url{https://git.io/ASGNet}.}
\end{abstract}

\section{Introduction}
\label{sec:intro}

Humans have a remarkable ability to learn how to recognize novel objects 
after seeing only a handful of exemplars. On the other hand, deep learning based computer vision systems have made tremendous progress, but have largely depended on large-scale training sets.
Also, 
deep networks mostly work with predefined classes and are incapable of generalizing to new ones.
The field of few-shot learning studies the development of such learning ability in artificial learning systems, where only a few examples of the new category are available.

In this work, we 
tackle the few-shot segmentation problem, where the target is learning to segment objects in
a given query image while only a few support images with ground-truth segmentation masks are available.
This is a challenging problem as the test data are novel categories which do not exist in the training set, and there are usually large variations in appearance and shape between the support and query images.



\begin{figure}[tbp]
\begin{center}
\includegraphics[width=8.5cm]{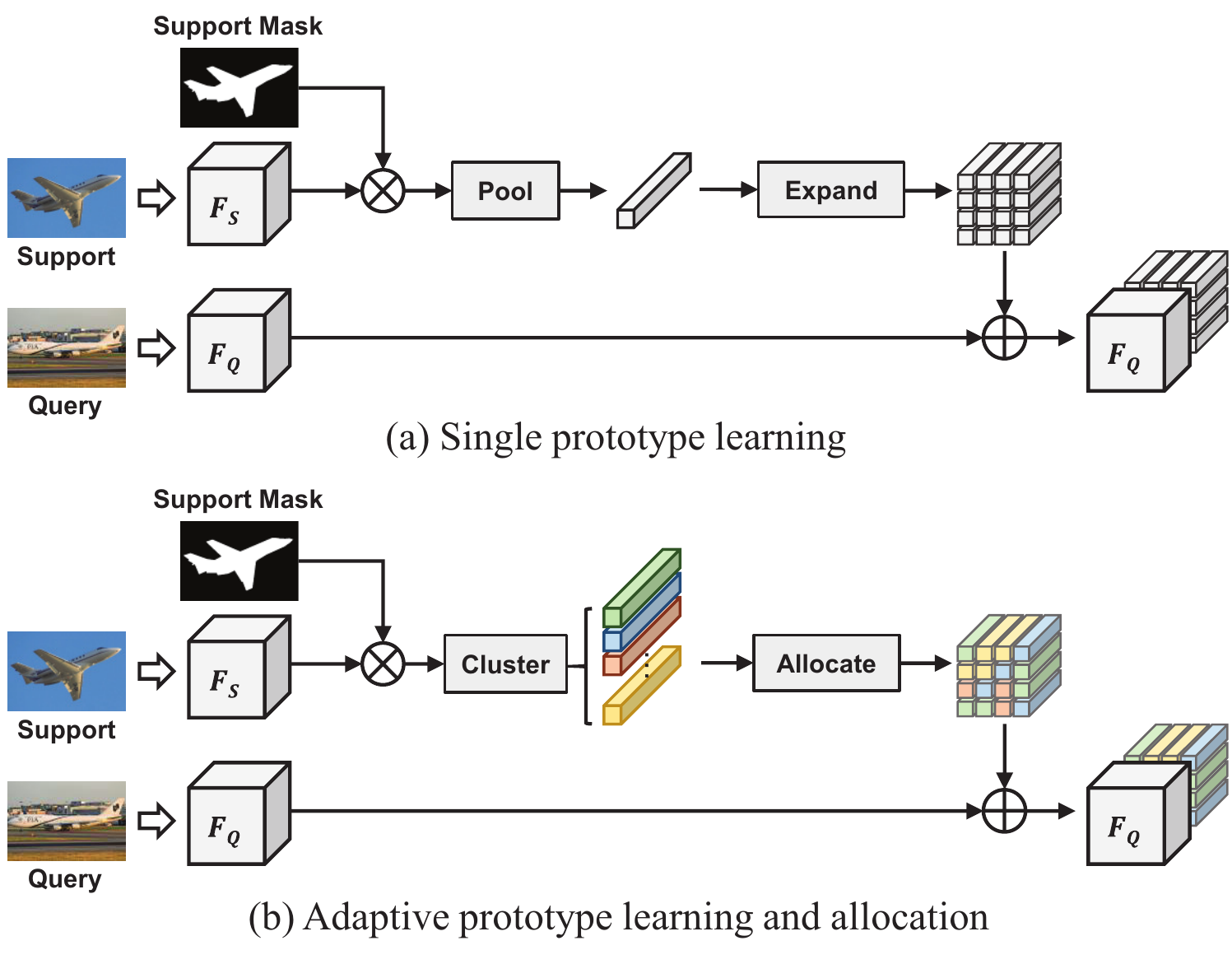}
\end{center}
\vspace{-0.3cm}
\caption{Comparison between (a) single prototype learning
and (b) proposed adaptive prototype learning and allocation. We utilize superpixel-guided clustering to generate multiple prototypes and then allocate them pixel-wise to query feature.}
\label{figure1}
\end{figure}

Current few-shot segmentation networks usually extract features from both query and support images,
and then propose different approaches for feature matching and object mask transfer 
from support to query image.
This feature matching and mask transfer are usually performed in one of two ways: prototypical feature learning or affinity learning.
Prototypical learning techniques condense the masked object features in a support image into a single or
few prototypical feature vectors. Then these techniques find the pixel locations of 
similar features in the query image to segment the desired object.
A key advantage of prototype learning is that the prototypical features are more robust to noise
than pixel features. However, prototypical features inevitably drop spatial information,
which is important when there is a large variation in the object appearance between the support and
query images. 
In addition, most prototypical learning networks~\cite{canet,sgone,panet,pfenet,pl} merely generate a single prototype by masked average pooling as shown in Figure \ref{figure1}(a), thus losing information as well as discriminability.

Affinity learning techniques~\cite{pgnet, brinet, danet} on the other hand, directly try to match
object pixels in a support image to query image pixels thereby transferring the object mask.
These techniques predict cross-image pixel affinities (also called connection strengths) using learned features, which perform feature matching while preserving spatial information better than
prototypical learning approaches. However, affinity learning techniques are prone to over-fitting on training data as they try to solve an under-constrained pixel-matching problem with dense affinity matrices.

In this work, we propose a novel prototypical learning technique that addresses some of the
main shortcomings of existing ones. In particular, we want to adaptively
change the number of prototypes and their spatial extent based on the image content, 
making the prototypes content-adaptive and spatially-aware. 
This adaptive, multi-prototype strategy is important to deal
with large variations in object scales and shapes across different images.
Intuitively, when an object occupies a large portion of the image, 
it carries more information and thus requires more prototypes to represent all the necessary information.
On the contrary, if the object is fairly small and 
the proportion of the background is large, then a single or few prototypes are sufficient.
In addition, we want the support region (spatial extent) for each of the
prototypes to be adaptive to object information that is present in the support image.
Concretely, we aim to divide the support feature into several representative areas according to the feature similarity.
We also want to adaptively choose more important prototypes while finding similar features in a query image.
As different object parts are visible in different image regions and in different
query images, we want to dynamically allocate different prototypes across query image 
for feature matching. For example, some parts of the object can be occluded
in a query image and we want to dynamically choose the prototypes that are corresponding to the visible parts in the query image.

We achieve this adaptive, multi-prototype learning and allocation with our 
Adaptive Superpixel-guided Network (ASGNet) that leverages superpixels
for adapting both the number and support regions of the prototypes. The schematic illustration is presented in Figure \ref{figure1}(b).
In particular, we propose two modules, Superpixel-guided Clustering (SGC)
and Guided Prototype Allocation (GPA), which form the core of ASGNet.
The SGC module does fast feature-based superpixel extraction on the
support image and the resulting superpixel centroids are considered
as prototypical features. Since superpixel shapes and numbers are adaptive
to the image content, the resulting prototypes also become adaptive.
The GPA module uses an attention-like mechanism to allocate most relevant support
prototype features to each pixel in a query image.
In summary, the SGC module provides adaptive prototype learning both in terms of the number of prototypes and their spatial extents, and the GPA module provides adaptive allocation of the learned prototypes when processing query features. 
These two modules make ASGNet highly flexible and adaptive to varying
object shapes and sizes, allowing it to generalize better to unseen object categories.
We make the following contributions:
\begin{itemize}
    \item We propose the Adaptive Superpixel-guided Network (ASGNet), 
    a flexible prototypical learning
    approach for few-shot segmentation 
    that is adaptive to different object scales, shapes and occlusions.
    \item We introduce two novel modules, namely
    Superpixel-guided Clustering (SGC) and Guided prototype allocation (GPA), for 
    adaptive prototype extraction and allocation respectively. They can serve as effective plug-and-play components on feature matching.
    \item ASGNet achieves top-performing results with fewer parameters and less computation. Specifically, the proposed method obtains mIoUs of 64.36\%/42.48\% in the 5-shot setting on Pascal-$5^i$/COCO-$20^i$, exceeding the state-of-the-art by 2.40\%/5.08\%.
\end{itemize}

\section{Related Work}
\label{sec:related_works}

\vspace{1mm}
\noindent \textbf{Semantic Segmentation.}
Most existing semantic segmentation methods are based on fully convolutional networks (FCNs)~\cite{fcn}, which replace fully connected layers with fully convolutional ones for pixel-level prediction.
Recent breakthroughs in semantic segmentation have mainly come from multi-scale feature aggregation~\cite{psp,deeplabv3+,denseaspp,adaptivepc} or attention mechanisms~\cite{dualatten,ocnet,emanet,hieratta,hamnet,ann,acfnet}. These methods often use dilated convolution kernels~\cite{dilated} and set up an encoder-decoder structure to obtain a large receptive field while maintaining the feature resolution.
Although these methods achieve tremendous success, they need long training time and a large amount of pixel-level labeled ground truth to fully supervise the network.
Also, in the inference stage, trained models cannot recognize new classes that do not exist in the training set.

\vspace{1mm}
\noindent \textbf{Few-shot Learning.}
Few-shot learning focuses on the generalization ability of models, 
so that they can learn to predict new classes given only a few annotated examples.
Existing methods mainly concentrate on metric-learning~\cite{matching,learning_to_compare,prototypical_net} and meta-learning~\cite{learning_to_learn,model_agnostic_ml,optimi}. 
The core idea of metric learning is distance measurement, and it is generally formulated as an optimization of distance/similarity between images or regions.
In meta-learning approaches, the main idea is to define specific optimization or loss functions to achieve fast learning capability.
Among these methods, the concept of prototypical networks has been extensively adopted in few-shot segmentation, which largely reduces computational budget while maintaining high performance.
Most methods focused on image classification, while recently few-shot segmentation has received growing attention.

\vspace{1mm}
\noindent \textbf{Few-shot Segmentation.}
Few-shot segmentation is an extension of few-shot classification, and it tackles a more challenging task of predicting a label at each pixel instead of predicting a single label for the entire image.
This research problem was introduced by Shaban \etal~\cite{oslsm}, who proposed a classical two-branch network.
Later on, PL~\cite{pl} introduced the idea of using prototypes. In that work, each prediction is generated by measuring the similarity between prototypes and pixels in query image.
SG-One~\cite{sgone} proposed masked average pooling to obtain the object-related prototype, which has been widely used in subsequent works.
PANet~\cite{panet} introduced a novel prototype alignment regularization to fully exploit the support knowledge, making the final prediction with only the measurement of cosine distance.
Based on masked average pooling, CANet \cite{canet} expanded the prototype to the same size of the query feature and concatenated them together. This work also used an iterative optimization module to refine the segmentation result.
PGNet, BriNet and DAN \cite{pgnet,brinet,danet} introduced dense pixel-to-pixel connections between support and query features to maintain spatial information.
More recently, PMMs~\cite{pmms} leveraged the expectation-maximization (EM) algorithm to generate multiple prototypes.
However, all prototypes have the same relevance in this model, which can potentially be sensitive to poorly matched prototypes. 
Instead, in our work we utilize the similarity between each prototype and the query feature to select the most relevant prototype at each pixel location.

\vspace{1mm}
\noindent \textbf{Superpixel Segmentation.}
A superpixel is defined as a set of pixels with similar characteristics (color, texture, category). Superpixels are effective in many computer vision tasks, and have recently been used as the basic units for few-shot segmentation \cite{ppnet, selfss}.
Superpixels carry more information than pixels, and can provide more compact and convenient image representations for the downstream vision tasks.
For more details about traditional and existing methods on superpixels, please refer to \cite{achanta2012slic,spsurvey}.

Our research is inspired by the maskSLIC \cite{maskslic} and superpixel sampling network (SSN) \cite{ssn}.
MaskSLIC adapts SLIC \cite{achanta2012slic} to a defined region of interest (RoI), and the main contribution is in the placement of seed points within the RoI.
SSN \cite{ssn} proposed the first end-to-end trainable superpixel algorithm by making the SLIC algorithm differentiable.
Inspired by the insights of these two techniques, we propose the masked superpixel clustering in feature space, which can gather similar features together and generate superpixel centroids as prototypes.
Instead of representing the information of the entire object, superpixel cenroids stand for the parts of the object with similar characteristics.

\section{Problem Definition}
The key difference between few-shot segmentation and a general semantic segmentation is that the categories in training and testing set do not intersect. 
It means that, in the inference stage, the testing set has classes totally unseen in the training. Specifically, given a training set $S_{train}=\{(I_{S/Q}, M_{S/Q})\}$ and testing set $S_{test}=\{(I_{S/Q}, M_{S/Q})\}$, the categories of the two sets do not intersect ($S_{train}\cap S_{test}=\emptyset$). Here $I\in\mathbb{R}^{H\times W\times 3}$ indicates the RGB image and $M\in\mathbb{R}^{H\times W}$ denotes the segmentation mask. Subscripts $S$ and $Q$ represent support and query, respectively.
Following the first one-shot segmentation work \cite{oslsm}, we align training and testing with the episodic paradigm \cite{matching}.
In each episode, the input to the model is composed by a query image $I_{Q}$ and $K$ samples $(I_{S}^i, M_{S}^i), i \in \{1,...,K\}$ from the support set. All support and query images have the same class $c$. We estimate the query mask $\tilde{M}_{Q}$ to approximate the ground truth mask $M_{Q}$.


\begin{figure*}[]
\centering
\includegraphics[width=16.5cm]{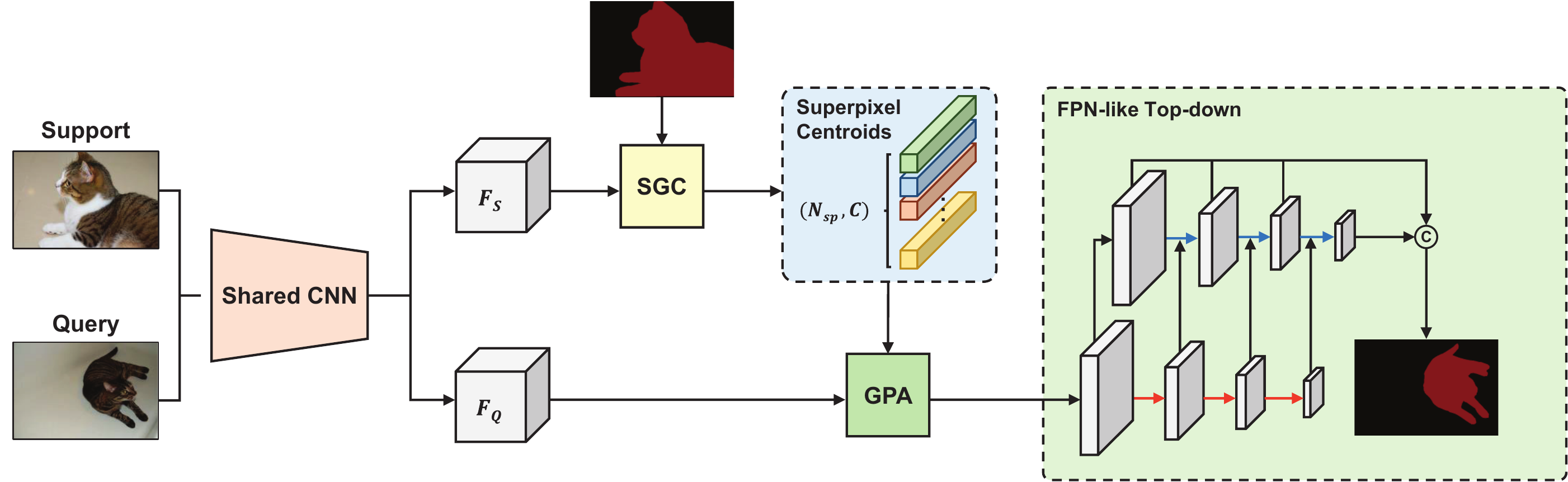}
\caption{Overall architecture of the proposed Adaptive Superpixel-guided Network. }
\label{overall}
\end{figure*}

\section{Proposed Method}
In this section, we first introduce the proposed two modules for prototype generation and matching, which are superpixel-guided clustering (SGC) and guided prototype allocation (GPA).
Then, we discuss the adaptive ability of these two modules. 
After that, we introduce the overall network architecture, named Adaptive Superpixel-guided Network (ASGNet), which integrates the SGC and GPA modules in one model.
The overall structure is shown in Figure \ref{overall}.
Finally, we elaborate the $k$-shot setting in ASGNet.

\subsection{Superpixel-guided Clustering}

The core idea of SGC is inspired by the superpixel sampling network (SSN) \cite{ssn} and MaskSLIC \cite{maskslic}.
SSN was the first end-to-end trainable deep network for superpixel segmentation. The key contribution in SSN is converting the nearest-neighbor operation in SLIC~\cite{achanta2012slic} into a differentiable one.
The traditional SLIC superpixel algorithm uses k-means clustering iteratively with two steps: \textit{Pixel-superpixel association} and \textit{Superpixel centroid update}.
Based on the color similarity and proximity, pixels are assigned to different superpixel centroids.
Specifically, the input image $I\in\mathbb{R}^{n\times 5}$ is usually in a five-dimensional space (labxy) with $n$ pixels, where lab represents the pixel vector in CIELAB color space and xy indicates the pixel location.
After iterative clustering, the algorithm outputs the association map where each pixel $n$ is assigned to one of the $m$ superpixels.

\begin{algorithm}[ht]
\caption{Superpixel-guided Clustering (SGC)} 
 {\bf Input:} 
Support feature $F_{s}$, support mask $M_{s}$, and initial superpixel seeds $S^0$
\begin{algorithmic}[0]
\State \textbf{concatenate} the absolute coordinates to $F_{s}$
\State \textbf{extract} masked features $F_{s}^{'}$ via support mask $M_{s}$
\For{each iteration $t$} 
    \State Compute association between each pixel $p$ and superpixel $i$, $Q_{pi}^{t}=e^{{-\left \| F_{p}^{'} - S_{i}^{t-1} \right \|}^{2}}$
    \State Update superpixel centroids, 
    \\
    $S_{i}^{t}=\frac{1}{Z_{i}^{t}}\sum _{p=1}^{N_{m}}Q_{pi}^{t}F_{p}^{'}$; $Z_i^t=\sum_{p}Q_{pi}^t$
\EndFor
\State \textbf{end for}
\State \textbf{remove} coordinates information
\\
\Return final superpixel centroids $S$ \Comment{$(N_{sp}, C)$}	
\end{algorithmic}
\label{alg1}
\end{algorithm}

This simple method inspires us with an insightful idea, which is to aggregate the feature map into multiple superpixel centroids in a clustering way, and here superpixel centroids can serve as prototypes.
Therefore, instead of computing the superpixel centroids in image space, we estimate them in feature space by clustering similar feature vectors.
The whole SGC process is delineated in Algorithm~\ref{alg1}.

Given support feature $F_s\in \mathbb{R}^{c\times h\times w}$, support mask $M_s\in \mathbb{R}^{h\times w}$ and initial superpixel seeds $S^{0}\in \mathbb{R}^{c\times N_{sp}}$ ($N_{sp}$ is the number of superpixels), we aim to obtain the final superpixel centroids, which act as multiple compact prototypes.
First, we concatenate the coordinates of each pixel with the support feature map to introduce positional information.
Then, we define the distance function $D$ following SLIC \cite{achanta2012slic}, which consists of feature and spatial distance:
\begin{equation}
	D =\sqrt{(d_f)^2 + (d_s/r)^2},
\end{equation}
where $d_f$, $d_s$ are the Euclidean distance for features and coordinate values, and $r$ is a weighting factor.
We filter out the background information with the support mask and only keep the masked features,
compressing the feature map from $F_s\in\mathbb{R}^{c\times h\times w}$ to  $F_s^{'} \in \mathbb{R}^{c\times N_{m}}$, where $N_{m}$ is the number of pixels inside the support mask.

\begin{figure}[t]
\centering
\includegraphics[width=8.4cm]{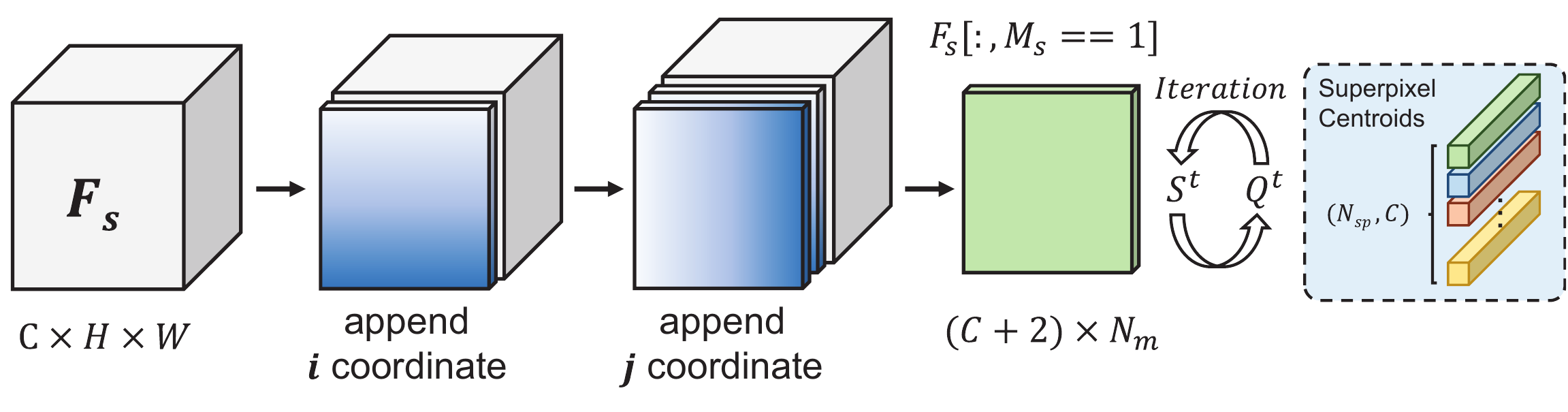}
\caption{Illustration of proposed Superpixel-guided Clustering.}
\label{SGC}
\end{figure}

\begin{figure*}[htbp]
\centering
\includegraphics[width=16.5cm]{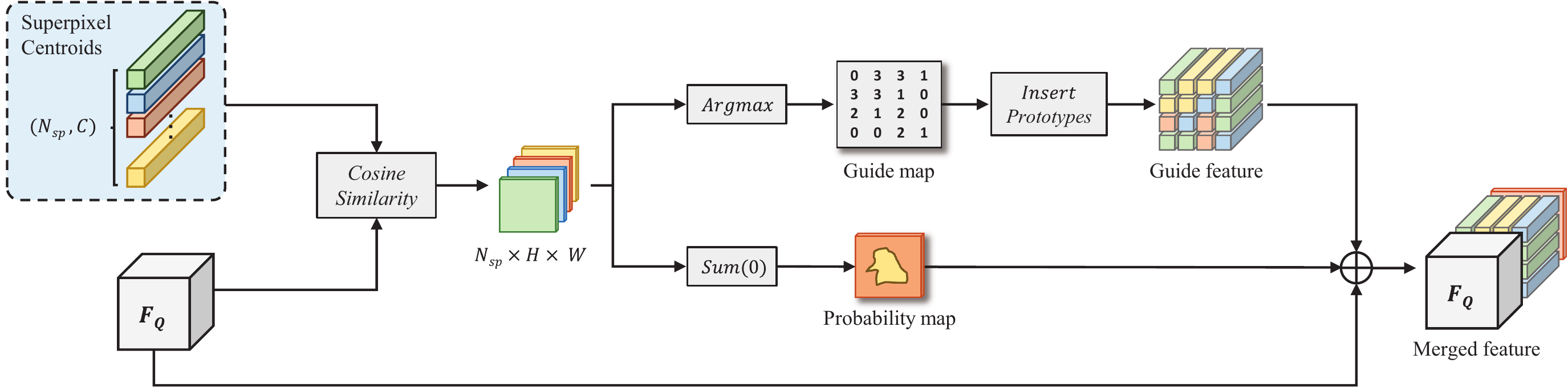}
\vspace{0.1cm}
\caption{Illustration of proposed Guided Prototype Allocation.}
\label{GPA}
\end{figure*}

\begin{figure}[htbp]
\centering
\includegraphics[width=8.4cm]{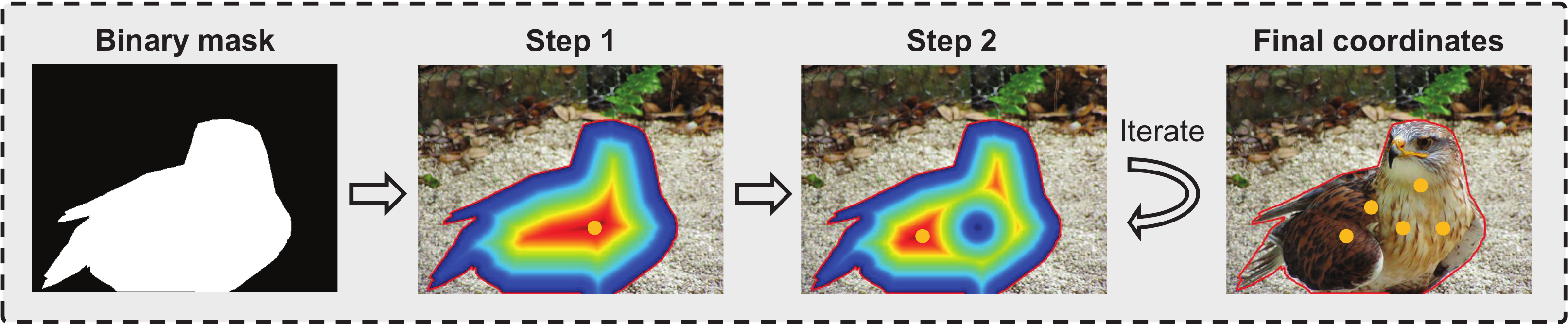}
\caption{The pipeline of placing initial seeds. (1) Step 1: We conduct Distance Transform and choose the maximum point. (2) Step 2: We set that point to False and then repeat Step 1. After that, we iterate Step 2 until we get coordinates of all initial seeds.}
\label{seed_coordinates}
\end{figure}

Then we compute superpixel-based prototypes in an iterative fashion. 
For each iteration $t$, we first compute the association map $Q^{t}$ between each pixel $p$ and all superpixels according to the distance function $D$:
\begin{equation}
Q_{pi}^{t}=e^{-D(F_{p}^{'}, S_i^{t-1})}=e^{{-\left \| F_{p}^{'} - S_{i}^{t-1} \right \|}^{2}}.
\end{equation}
Then, new superpixel centroids are updated as the weighted sum of masked features:
\begin{equation}
S_{i}^{t}=\frac{1}{Z_{i}^{t}}\sum _{p=1}^{N_{m}}Q_{pi}^{t}F_{p}^{'},
\end{equation}
where  $Z_i^t=\sum_{p}Q_{pi}^t$ is a constant for normalization.
The above process is visualized in Figure \ref{SGC}. 

Here, we elaborate the selection of initial seeds.
Generally, in superpixel algorithm, a $H\times W$ image is evenly partitioned into regular grid cells of size $h\times w$, and each grid cell is considered as an initial seed (i.e., superpixel).
However, this initialization is not suitable for our purposes where a foreground mask is given for the support image
and we only need to initialize the seeds inside this foreground region.
To uniformly initialize seeds
in the masked region, we refer to MaskSLIC~\cite{maskslic} for iteratively placing each initial seed, and the pipeline is depicted in Figure \ref{seed_coordinates}.
This seed initialization results in faster convergence of superpixel-guided clustering with only a few iterations.


\subsection{Guided Prototype Allocation}
\begin{figure*}[]
\centering
\includegraphics[width=16.5cm]{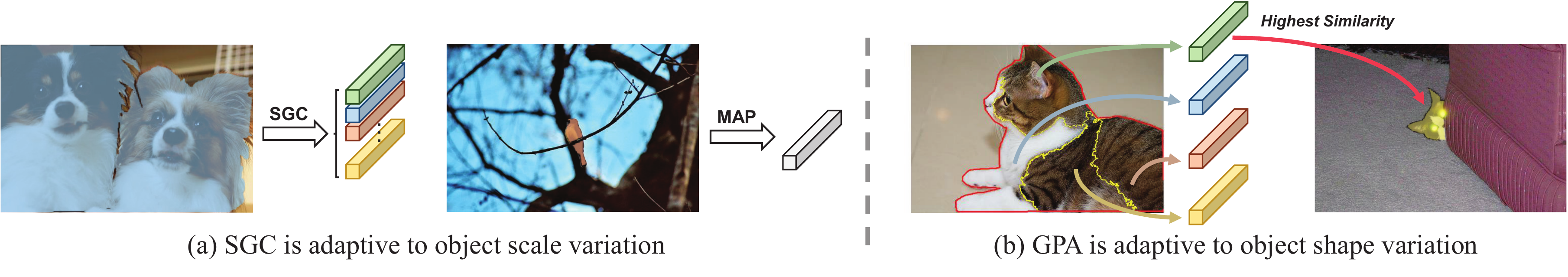}
\caption{Illustration of the model adaptability in prototype learning and allocation. SGC and GPA indicate the proposed superpixel-guided clustering and guided prototype allocation respectively, and MAP denotes masked average pooling proposed in \cite{sgone}.}
\label{Adaptability}
\end{figure*}

After extracting prototypes, previous methods mostly follow the design of CANet~\cite{canet}, expanding a single prototype to the same size of the query feature and concatenating them together.
However, this operation 
results in equal guidance to every location in the query feature.
To make prototype matching more adaptive to the query image content, we propose the Guided Prototype Allocation (GPA), illustrated in Figure~\ref{GPA}.
We first compute the cosine distance to measure the similarity between each prototype and each query feature element:
\begin{equation}
C_{i}^{x, y}=\frac{S_{i}\cdot F^{x,y}_q}{\left \| S_{i}\right \|\cdot \left \| F^{x,y}_{q}\right \|}  \quad  i\in \{ 1, 2, ..., N_{sp}\},
\end{equation}
where $S_i\in \mathbb{R}^{c\times 1}$ is the $i_{th}$ superpixel centroid (prototype), and $F^{x,y}_q\in \mathbb{R}^{c\times 1}$ is the feature vector at location $(x, y)$ of the query feature.
We use this similarity information as input to a two-branch structure. The first branch computes which prototype is the most similar at each pixel location as follows:
\begin{equation}
G^{x, y}=\mathop{\arg \max}_{i\in\{0,..,N_{sp}\}}C_{i}^{x, y},
\end{equation}
where the argmax operator is adopted to obtain $G^{x, y}$, which is a single index value representing a particular prototype.
Putting all index values together, we get a guide map $G\in \mathbb{R}^{h\times w}$.
Then, by placing corresponding prototype in each position of the guide map, we obtain the guide feature $F_{G}\in \mathbb{R}^{c\times h\times w}$ to achieve pixel-wise guidance.
While in the other branch, all the similarity information $C$ is added up 
across all the superpixels to get the probability map $P$.

Finally, we concatenate the probability map and the guide feature with the original query feature $F_Q$ to provide the guiding information, and thus obtain the refined query feature $F_Q^{'}$:
\begin{equation}
{F}_{Q}^{'} = f(F_{Q} \oplus F_{G} \oplus P),
\end{equation}
where $\oplus$ indicates the concatenation operation along channel dimension, and $f(\cdot)$ is a $1\times 1$ convolution.

\subsection{Adaptability}
\label{sec:adaptability}
As mentioned before, we argue that one of the key attributes in the proposed network is its adaptive ability for few-shot semantic segmentation.
In Figure~\ref{Adaptability}, we provide some examples illustrating the adaptive ability of SGC and GPA.
In SGC, to make it adaptive to object scale, we define a criterion to regulate the number of superpixel centroids as
\begin{equation}
N_{sp} = \min(\left \lfloor \frac{N_{m}}{S_{sp}}\right \rfloor, N_{max}),
\label{eqn:adaptability}
\end{equation}
where $N_{m}$ is the number of pixels in the support mask; $S_{sp}$ is the average area assigned to each initial superpixel seed, and we set to 100 empirically.
When the foreground is fairly small, $N_{sp}=0$ or $1$, this method degrades to a general masked average pooling as shown in Figure \ref{Adaptability}(a).
In addition, to reduce the computational burden, we set a hyper-parameter $N_{max}$ to constrain the maximum number of prototypes.
In GPA, we can observe its adaptability to object shape. 
In other words, it is resilient to occlusion.
When severe occlusion exists in query image, e.g., Figure \ref{Adaptability}(b), GPA can choose the best matched prototype for each query feature location.

\subsection{Adaptive Superpixel-guided Network}
Based on the above SGC and GPA modules, we propose the Adaptive Superpixel-guided Network (ASGNet) for few-shot semantic segmentation as illustrated in Figure \ref{overall}.
First, the support and query images are fed into a shared CNN (pretrained on ImageNet \cite{imagenet}) to extract features.
Then, by passing the support features through SGC with support mask, we obtain the superpixel centroids, 
which are considered as prototypes.
After that, for more accurate pixel-wise guidance, we adopt GPA module 
to match the prototypes with the query feature.
Finally, we use the feature enrichment module~\cite{pfenet} and set up a FPN-like \cite{fpn} top-down structure to introduce multi-scale information.
As demonstrated in \cite{pfenet}, transferring features from fine to coarse promotes the feature interaction, so we follow their design for fast multi-scale aggregation.
Finally, all different scales are concatenated, and each scale yields a segmentation result for computing the 
loss.

\subsection{K-shot Setting}
\label{sec:k-shot}
In previous work, the $k$-shot setting is 
usually tackled via feature averaging or attention-based fusion.
However, it turns out that the improvements from such 
strategies is minor, while requiring heavy computation.
In contrast, based on the proposed SGC and GPA, our ASGNet can easily 
adopt 
an efficient $k$-shot strategy without collapsing support features.
Specifically, in each support image and mask pair, we implement SGC to obtain the superpixel centroids.
By collecting all superpixel centroids together, we get the overall superpixel centroids $S$ from $k$ shots:
\begin{equation}
S = (S^1, S^2, ..., S^k),
\end{equation}
where the number of superpixel centroids is $N_{sp}=\mathop{\sum}^{k}_{i=1}{N_{sp}^i}$.
By doing so, the GPA can receive a larger range of selections from multiple shots, and thus yield more 
accurate guidance to segment the object in the query image.

\section{Experiments}
\subsection{Datasets and Evaluation Metric}
We choose Pascal-$5^i$ and COCO-$20^{i}$, two widely used datasets for few-shot semantic segmentation, to analyze our model performance.
Pascal-$5^i$~
\cite{oslsm} 
includes images from the PASCAL VOC 2012 \cite{pascal} and extra annotations from SBD \cite{sbd}.
The total 20 categories are evenly partitioned into four splits, and the model training is conducted in a cross-validation manner.
Specifically, three splits are selected in the training process and the remaining one is used for testing.
During inference, 1000 support-query pairs are randomly sampled for evaluation~
\cite{oslsm}.
Different from Pascal-$5^i$, MSCOCO \cite{mscoco} is a large-scale dataset with 82,081 images 
in the training set.
Following FWB \cite{fwb}, the overall 80 classes from MSCOCO~\cite{mscoco} are evenly divided into four splits with the same cross-validation strategy.
For more stable results, we randomly sample 20,000 pairs during evaluation~\cite{pfenet}.

We use mean intersection-over-union (mIoU) 
as a primary evaluation metric for the ablation study, as it is commonly recognized in image segmentation.
In addition, for more consistent comparisons, results of foreground-background IoU (FB-IoU) are also reported.

\subsection{Implementation Details}
We adopt ResNet \cite{resnet} as the backbone network~\cite{canet},
and we concatenate block2 and block3 to generate feature maps.
We train the model 
with a SGD optimizer on Pascal-$5^{i}$ for 200 epochs and COCO-$20^i$ for 50 epochs.
We set the initial learning rate to 0.0025 with batch size 4 on Pascal-$5^i$, and 0.005 with batch size 8 on COCO-$20^i$.
The number of iterations in the SGC module is set to 10 during training and 5 during inference. 
We use data augmentation during training - input images are transformed with random scale, horizontal flip and rotation from [-10, 10], and then all images are cropped to $473 \times 473$ (Pascal) or $641 \times 641$ (COCO) as training samples.
We implement our model using Pytorch and run our experiments on a workstation with Nvidia Tesla V100 GPU.
To increase the variance of the cosine measurement, we remove the ReLU layer before both support and query features, so the similarity metric is bounded in [-1, 1] rather than [0, 1].

\subsection{Ablation Study}
To verify the effectiveness of the proposed modules, we implement extensive ablation studies with a 
ResNet-50 backbone on Pascal-$5^i$.
We use floating point operations (FLOPs) to represent the amount of computation, and take both addition and multiplication into account.

\label{max_proto}
\vspace{1mm}
\noindent{\textbf{Number of Superpixel Centroids.}}
To explore the effect of the number of superpixel centroids, we conduct experiments with different $N_{max}$ in 1-shot segmentation.
As stated in Section \ref{sec:adaptability}, $N_{max}$ is a hyper-parameter set to regulate the maximum number of prototypes.
Particularly, when $N_{max}=1$, the prototype generation process degrades to the masked average pooling.
As shown in Table \ref{proto_number}, when $N_{max}$ equals 5, ASGNet achieves the best results in split 0 and 2, as well as the best mean performance, which 
demonstrates the validity of superpixel-guided clustering.
The results get improved gradually when $N_{max}$ increases from 1 to 5, and then decrease slightly after that, denoting that excessive prototypes bring negative effect and are prone to over-fitting.
Finally, we choose $N_{max}$ as 5 for both 1-shot and 5-shot segmentation.
In addition, we analyzed
the impact of the proposed adaptive scheme on the number of superpixel centroids $N_{sp}$ and show the results in Table~\ref{adaptive_proto}.
Compared with using the fixed number (5) of superpixels, the results show that the adaptive setting (Eqn.~\ref{eqn:adaptability}) can reduce redundant computation while obtaining 
better performance.



\begin{table}[htbp]
\begin{center}
\begin{tabular}{|c|cccc|c|}
\hline
$N_{max}$ & s-0 & s-1 & s-2 & s-3 & mean\\ \hline \hline
1        &  57.40       & 67.52       &  55.69     & \textbf{54.36}    & 58.74 \\
3        &  57.92       & 67.86       &  55.86     & 53.74    & 58.85 \\
5        &  \textbf{58.84}       & 67.86       &  \textbf{56.79}     & 53.66    & \textbf{59.29} \\ 
7        &  \textbf{58.84}       & \textbf{68.11}       &  55.85     & 54.00    & 59.20 \\ 
9        &  57.61       & 67.97       &  56.68     & 53.27    & 58.88 \\ \hline
\end{tabular}
\end{center}
\caption{Ablation study on the maximum number of superpixel centroids $N_{max}$. s-$x$ denotes different cross-validation splits.}
\label{proto_number}
\end{table}

\begin{table}[htbp]
\begin{center}
\begin{tabular}{|c|cccc|c|}
\hline
Centroids & s-0 & s-1 & s-2 & s-3 & mean\\ \hline \hline
Fixed       &  57.87       & 67.62       &  \textbf{57.10}     & 53.61    & 59.05 \\
Adaptive & \textbf{58.84}       & \textbf{67.86}       &  56.79     & \textbf{53.66}    & \textbf{59.29} \\ \hline
\end{tabular}
\end{center}
\caption{Performance comparison between fixed and adaptive superpixel centroids.}
\label{adaptive_proto}
\end{table}

\vspace{1mm}
\noindent{\textbf{SGC and GPA.}}
To demonstrate the effectiveness of proposed SGC and GPA modules, we conduct diagnostic experiments on prototype generation and matching.
We first implement a baseline model with the single prototype learning proposed in PFENet \cite{pfenet}.
Then, we introduce SGC to generate multiple prototypes, and fuse them in a dense 
manner with reference to PMMs \cite{pmms}.
Finally, we replace the prototype expansion with the proposed allocation scheme.
For ablations on the SGC module, as shown in Table \ref{methods_ablation}, in the 1-shot setting we observe that replacing mean average pooling (MAP) with SGC can worsen the representations and lead to degraded performance, but the model benefits in the 5-shot scenario.
We deem the main reason is that excessive prototypes become highly similar on a single support sample, and cosine distance can not distinguish them apart.
Finally, when GPA module is adopted for prototype matching, the performance improves by 2.70\% compared to the prototype expansion, and also the computational overhead is
much lower.

\begin{table}[htbp]
\begin{center}
\resizebox{84.3mm}{10.9mm}{
\begin{tabular}{|cc|cc|cc|c|}
\hline
\multicolumn{2}{|c|}{Generation}              & \multicolumn{2}{c|}{Matching}              & \multicolumn{2}{c|}{mIoU} & \multicolumn{1}{c|}{\multirow{2}{*}{FLOPs$\Delta$}} \\ \cline{1-6}
\footnotesize{MAP}   & \footnotesize{SGC}  & \footnotesize{Expand}   & \footnotesize{GPA}  & 1-shot & 5-shot & \multicolumn{1}{c|}{}                       \\ \hline \hline
        \checkmark  &           &   \checkmark      &               &  58.96    &  60.19       & 0.9G              \\
         &  \checkmark          &   \checkmark      &               &  58.31     & 61.24       & K*8.5G+0.5G             \\
  &       \checkmark    &         &  \checkmark       & \textbf{59.29}  & \textbf{63.94}    & K*5.5M+0.9G          \\ \hline
\end{tabular}}
\end{center}
\caption{Ablation study on prototype generation (Masked average pooling (MAP) vs. SGC) and matching (Expand vs. GPA). FLOPs$\Delta$ denotes the computational cost from prototype matching process, and K is the adaptive number of prototype (K$\le5$).}
\label{methods_ablation}
\end{table}

\vspace{1mm}
\noindent{\textbf{K-shot Fusion Setting.}}
As mentioned in Section \ref{sec:k-shot}, ASGNet is able to tackle the $k$-shot learning problem without collapsing the support features.
To explore the effect of different fusion methods, we compare our $k$-shot setting with two other commonly used solutions: 1) feature average fusion \cite{co-fcn} and 2) attention weighted summation \cite{canet}. 
For simplicity, experiments of this ablation study are only conducted in Pascal-$5^{0}$.
As reported in Table \ref{k-shot}, our simple strategy achieves the best performance and the largest increment over the 1-shot baseline (4.82\%) without additional computation.
On the contrary, attention-based fusion requires a large amount of computation, and has limited performance improvement.
This demonstrates that the GPA module is highly effective when given a large number of selections.

\begin{table}[htbp]
\begin{center}
\begin{tabular}{|l|ccc|}
\hline
5-shot setting & mIoU & IoU$\Delta$ & FLOPs$\Delta$ \\ \hline \hline
1-shot basline &     58.84     &  -   & -      \\ \hline
Attention    & 60.69         & 1.85 & 42.6G      \\ \hline
Feature-avg     &  62.10        &  3.26   & 3.7M      \\ \hline
Ours      & \textbf{63.66}       & \textbf{4.82}    &  None     \\ \hline
\end{tabular}
\end{center}
\caption{Ablation study of different $k$-shot fusion strategy.}
\label{k-shot}
\end{table}

\subsection{Comparison to State-of-the-art}

\begin{table*}[htbp] \small
\begin{center}
\begin{tabular}{|l|l|ccccc|ccccc|c|}
\hline
\multicolumn{1}{|l|}{\multirow{2}{*}{\textbf{Backbone}}} & \multirow{2}{*}{\textbf{Methods}} & \multicolumn{5}{c|}{\textbf{1-shot}}                                                                              & \multicolumn{5}{c|}{\textbf{5-shot}}                                                                             & \multirow{2}{*}{$\Delta$} \multirow{2}{*}{}         \\ \cline{3-12}
\multicolumn{1}{|l|}{}                          &                          & s-0            & s-1            & s-2            & s-3            & mean                                & s-0            & s-1            & s-2            & s-3            & mean                                &                           \\ \hline\hline
\multirow{6}{*}{VGG-16}                         & OSLSM \cite{oslsm}                    & 33.60          & 55.30          & 40.90          & 33.50          & 40.80                               & 35.90          & 58.10          & 42.70          & 39.10          & 43.95                               & 3.15                      \\
                                                & co-FCN \cite{co-fcn}                   & 36.70          & 50.60          & 44.90          & 32.40          & 41.10                               & 37.50          & 50.00          & 44.10          & 33.90          & 41.40                               & 0.30                      \\
                                                & AMP \cite{amp}                      & 41.90          & 50.20          & 46.70          & 34.40          & 43.40                               & 40.30          & 55.30          & 49.90          & 40.10          & 46.40                               & 3.00                      \\
                                                & SG-One \cite{sgone}                   & 40.20          & 58.40          & 48.40          & 38.40          & 46.30                               & 41.90          & 58.60          & 48.60          & 39.40          & 47.10                               & 0.80                      \\
                                                & PANet \cite{panet}                    & 42.30          & 58.00          & 51.10          & 41.20          & 48.10                               & 51.80          & 64.60          & 59.80          & 46.50          & 55.70                               & \textbf{7.60}                      \\
                                                & FWB \cite{fwb}                      & 47.04          & 59.64          & 52.61          & 48.27          & 51.90                               & 50.87          & 62.86          & 56.48          & 50.09          & 55.08                               & 3.18                      \\ \hline
\multirow{7}{*}{ResNet50}                       & CANet$\dagger$ \cite{canet}                    & 52.50          & 65.90          & 51.30          & 51.90          & 55.40                               & 55.50          & 67.80          & 51.90          & 53.20          & 57.10                               & 1.70                      \\
                                                & PGNet$\dagger$ \cite{pgnet}                    & 56.00          & 66.90          & 50.60          & 50.40          & 56.00                               & 57.70          & 68.70          & 52.90          & 54.60          & 58.50                               & 2.50                      \\
                                                & RPMMs \cite{pmms}                     & 55.15          & 66.91          & 52.61          & 50.68          & 56.34                               & 56.28          & 67.34          & 54.52          & 51.00          & 57.30                               & 0.96                      \\
                                                & SimPropNet \cite{simprop}               & 54.82          & 67.33          & 54.52          & 52.02          & 57.19                               & 57.20          & 68.50          & 58.40          & 56.05          & 60.04                               & 2.85                      \\
& PPNet \cite{ppnet}       & 47.83          & 58.75 & 53.80 & 45.63          & 51.50                               & 58.39          & 67.83          & \textbf{64.88}          & 56.73          & 61.96                               & \textbf{10.46} \\
 & PFENet \cite{pfenet}       & \textbf{61.70}          & \textbf{69.50} & 55.40 & \textbf{56.30}          & \textbf{60.80}                               & 63.10          & \textbf{70.70}          & 55.80          & \textbf{57.90}          & 61.90                               & 1.10 \\
 & ASGNet (ours) & 58.84 & 67.86          & \textbf{56.79}         & 53.66  & 59.29 & \textbf{63.66} & 70.55 & 64.17 & 57.38 & \textbf{63.94} & 4.65                      \\ \hline
\multirow{4}{*}{ResNet101}      & FWB \cite{fwb}                    & 51.30          & 64.49 & 56.71 & 52.24          & 56.19                               & 54.84          & 67.38          & 62.16          & 55.30          & 59.92                               & 3.73 \\
& DAN$\dagger$ \cite{danet}                    & 54.70          & 68.60 & \textbf{57.80} & 51.60          & 58.20                               & 57.90          & 69.00          & 60.10          & 54.90          & 60.50                               & 2.30 \\
& PFENet \cite{pfenet}       & \textbf{60.50}          & \textbf{69.40} & 54.40 & \textbf{55.90}          & \textbf{60.10}                    & 62.80          & 70.40          & 54.90          & \textbf{57.60}          & 61.40                               &  1.30 \\
& ASGNet (ours)  & 59.84 & 67.43          & 55.59          & 54.39 & 59.31 & \textbf{64.55} & \textbf{71.32} & \textbf{64.24} & 57.33 & \textbf{64.36} & \textbf{5.05}                      \\ \hline

\end{tabular}
\end{center}
\caption{Comparison with state-of-the-art on Pascal-$5^{i}$. $\dagger$ indicates multi-scale inference is adopted. $\Delta$ means increment over 1-shot segmentation result.}
\vspace{-3mm}
\label{pascal}
\end{table*}

\begin{table}[htbp] \small
\begin{center}
\begin{tabular}{|l|cc|c|c|}
\hline
\multirow{2}{*}{\textbf{Methods}} & \multicolumn{2}{c|}{\textbf{FB-IoU}} & \multirow{2}{*}{$\Delta$} & \multirow{2}{*}{\textbf{\#Params}} \\ \cline{2-3}
                                  & 1-shot             & 5-shot         &                   &                                    \\ \hline\hline
OSLSM \cite{oslsm}                            & 61.3               & 61.5           & 0.2               & 272.6M                             \\
co-FCN \cite{co-fcn}                           & 60.1               & 60.2           & 0.1               & 34.2M                              \\
AMP \cite{amp}                              & 62.2               & 63.8           & 1.6               & 14.9M                              \\
SG-One \cite{sgone}                            & 63.1               & 65.9           & 2.8               & 19.0M                              \\
CANet$\dagger$ \cite{canet}                            & 66.2               & 69.6           & 3.4               & 10.5M                              \\
PGNet$\dagger$ \cite{pgnet}                           & 69.9               & 70.5           & 0.6               & -                             \\
PANet \cite{panet}                            & 66.5               & 70.7           & 4.2               & 14.7M                              \\ 
DAN$\dagger$ \cite{danet}                            & 71.9               & 72.3           & 0.4               & -      \\
SimPropNet \cite{simprop}                            & 73.0               & 72.9           & -0.1               & -                              \\
PFENet \cite{pfenet}                            & \textbf{73.3}               & 73.9           & 0.6               & 10.8M                              \\ \hline
ASGNet (RN-50)                            & 69.2          & 74.2               &  \textbf{5.0}                 & \textbf{10.4M}    \\
ASGNet (RN-101)                            & 71.7          & \textbf{75.2}               & 3.5                 & \textbf{10.4M}     \\ \hline
\end{tabular}
\end{center}
\caption{Comparison of FB-IoU and the number of trainable parameters on Pascal-$5^i$. $\dagger$ indicates multi-scale inference is adopted. $\Delta$ means increment over 1-shot segmentation result.}
\vspace{-3mm}
\label{pascal_fb}
\end{table}

\begin{table}[htbp]\footnotesize
\begin{center}
\begin{tabular}{|l|l|cc|cc|}
\hline
\multirow{2}{*}{\textbf{Backbone}} & \multirow{2}{*}{\textbf{Methods}} & \multicolumn{2}{c|}{\textbf{mIoU}} & \multicolumn{2}{c|}{\textbf{FB-IoU}} \\ \cline{3-6} 
                                   &                                   & 1-shot               & 5-shot      & 1-shot                & 5-shot       \\ \hline\hline
\multirow{3}{*}{ResNet101}         & FWB \cite{fwb}                              & 21.19                & 23.65       & -                     & -            \\ 
                                   & DAN \cite{danet}                             & 24.20                & 29.60       & \textbf{62.30}        & 63.90        \\ 
                                   & PFENet \cite{pfenet}                             & 32.40                & 37.40       & 58.60        & 61.90        \\ \hline
\multirow{2}{*}{ResNet50}          & RPMMs \cite{pmms}                             & 30.58                & 35.52       & -                     & -            \\ 
                                   & ASGNet                            & \textbf{34.56}       &  \textbf{42.48}           & 60.39                 &  \textbf{66.96}            \\ \hline
\end{tabular}
\end{center}
\caption{Comparison with state-of-the-arts on COCO-$20^i$.}
\vspace{-5mm}
\label{coco}
\end{table}

\begin{figure}[t]
\centering
\includegraphics[width=8.4cm]{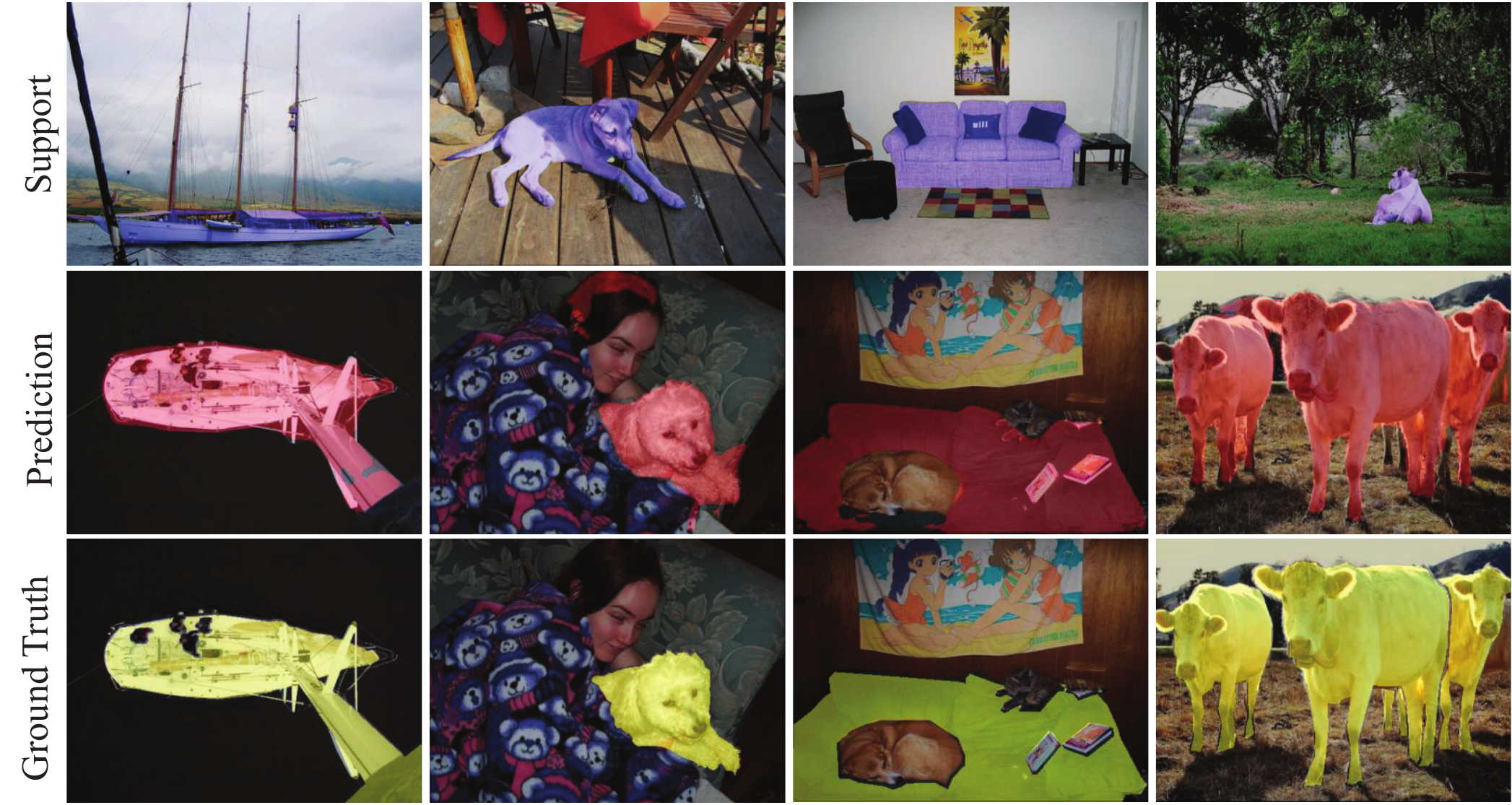}
\caption{Qualitative examples of 1-shot segmentation on the Pascal-$5^i$. Best viewed in color and zoom in.}
\vspace{-3mm}
\label{qualitative}
\end{figure}

\vspace{1mm}
\noindent{\textbf{Pascal-\bm{$5^{i}$}.}}
Comparisons to state-of-the-art methods are shown in Table \ref{pascal} and \ref{pascal_fb}, where two different metrics are adopted.
In Table \ref{pascal}, with ResNet-101 as the backbone, ASGNet outperforms recent methods with a considerable margin of 2.40\% in 5-shot segmentation, while being on par with state-of-the-arts under the 1-shot setting.
In Table \ref{pascal_fb}, we further make comparisons on FB-IoU and the number of trainable parameters.
Once again, the proposed ASGNet achieves significant improvement over state-of-the-arts in 5-shot setting (75.2\% vs.73.9\%), and it also has the largest performance increment of 5.0\% over 1-shot result.
In addition, we have far fewer trainable parameters than other methods.
In Figure \ref{qualitative}, we show some representative segmentation examples.
We observe that the proposed ASGNet can generate accurate segmentation results even when there are large variations in appearance and pose between support and query images.

\vspace{1mm}
\noindent{\textbf{COCO-\bm{$20^{i}$}.}} In Table \ref{coco}, we present the performance comparison of mean IoU and FB-IoU on COCO-$20^i$.
As can be seen, ASGNet achieves state-of-the-art results in both 1-shot and 5-shot setting in terms of the mean IoU, and it substantially outperforms previous methods.
Particularly, ASGNet achieves a margin of 3.98\% and 6.96\% higher mean IoU over RPMMs~\cite{pmms} in 1-shot and 5-shot segmentation respectively.
Also, our ASGNet obtains competitive 1-shot results and top-performing 5-shot results with respect to FB-IoU.
These results demonstrate that the proposed method is capable of handling more complex cases, as MSCOCO is a much more challenging dataset with diverse samples and categories.
\section{Conclusions}
In this paper we propose ASGNet for few-shot image segmentation.
Targeting the 
limitations of existing single prototype based models, we introduce two new modules, named Superpixel-guided Clustering (SGC) and Guided Prototype Allocation (GPA), for adaptive prototype learning and allocation.
Concretely, SGC aggregates similar feature vectors with feature-based superpixel clustering,
and GPA aims to allocate the most relevant prototype to each query feature element by measuring the similarity with cosine distance.
Extensive experiments and ablation studies have demonstrated the superiority of ASGNet, and we achieve state-of-the-art performance on both Pascal-$5^i$ and COCO-$20^i$ without any additional post-processing steps.

\vspace{1mm}
\noindent{\textbf{Acknowledgements}}
This work was supported by the National Research Foundation of Korea (NRF) grant funded by the Korea government (MSIT) (No.2020R1A2C101215911).

{\small

\bibliographystyle{ieee_fullname}
}
\clearpage



\appendix
\renewcommand\thesection{\Alph{section}}

\pagestyle{empty}
\thispagestyle{empty}

\begin{center}
\section*{Supplementary Material}    
\end{center}

Thank you for reading the supplementary material, in which we introduce more experimental details in Section \ref{1}, and provide more qualitative visualization examples on Pascal-$5^i$ and COCO-$20^i$ in Section \ref{2}.
\vspace{3mm}

\section{Additional Experimental Details}
\label{1}

\subsection{Detailed mean IoU results on COCO-\bm{$20^i$}}

In Table \ref{t1}, we present the detailed per-split results in terms of mean IoU.
As can be seen in the table, we achieve the best performance in every split, which demonstrates the superiority of our method.

\begin{table*}[htbp] \small
\begin{center}
\begin{tabular}{|l|l|ccccc|ccccc|}
\hline
\multirow{2}{*}{\textbf{Backbone}} & \multirow{2}{*}{\textbf{Methods}} & \multicolumn{5}{c|}{\textbf{1-shot}}                                               & \multicolumn{5}{c|}{\textbf{5-shot}}                                               \\ \cline{3-12} 
                                   &                                   & s-0            & s-1            & s-2            & s-3            & mean           & s-0            & s-1            & s-2            & s-3            & mean           \\ \hline \hline

\multirow{3}{*}{ResNet101}         & FWB                               & 16.98          & 17.98          & 20.96          & 28.85          & 21.19          & 19.13          & 21.46          & 23.93          & 30.08          & 23.65          \\
& DAN                               & -          & -          & -       & -          & 24.20          & -          & -          & -         & -          & 29.60          \\
& PFENet                               & 34.30        &33.00        &32.30       & 30.10         &32.40       &38.50       & 38.60      & 38.20        & 34.30         &37.40          \\
\hline
\multirow{2}{*}{ResNet50}          & RPMMs                              & 29.53          & 36.82          & 28.94          & 27.02          & 30.58          & 33.82          & 41.96          & 32.99          & 33.33          & 35.52          \\
                                   & ASGNet                            & \textbf{34.89} & \textbf{36.94} & \textbf{34.33} & \textbf{32.08} & \textbf{34.56} & \textbf{40.99} & \textbf{48.28} & \textbf{40.10} & \textbf{40.54} & \textbf{42.48} \\ \hline
\end{tabular}
\end{center}
\caption{Comparison with state-of-the-arts on COCO-$20^i$ with per-split results.}
\label{t1}
\end{table*}

\subsection{Calculation of FLOPs}
In the ablation study, we use floating point operations (FLOPs) to evaluate the amount of computation and model complexity.
Here, we describe the calculation in detail.
For a general convolution layer, the operations of one pixel in the output feature map are calculated as follows:
\begin{equation} \small
    F=\begin{cases}
 (C_{in} \cdot K^2)+  (C_{in} \cdot K^2 - 1) & \text{bias=False}\\ 
 (C_{in} \cdot K^2) +  (C_{in} \cdot K^2) & \text{bias=True}
\end{cases}
\end{equation}
where the first item is multiplication, and the second one denotes addition.
$C_{in}$ is the number of input channels and $K$ is the kernel size.
Then, extending to the whole feature map, we get the number of FLOPs as:
\begin{equation}
    FLOPs = F\times H\times W\times C_{out},
\end{equation}
where $H$, $W$ is the size of output feature, and $C_{out}$ is the number of output feature channels.
For example, the FLOPs are 0.9G when using 256 $1\times 1$ convolution filters to process the merged feature $F_{Q}^{'} \in \mathbb{R}^{513\times 60 \times 60}$.

\subsection{Ablation Study on Iteration Number}
\begin{figure}[htbp]
\begin{center}
\includegraphics[width=7.5cm]{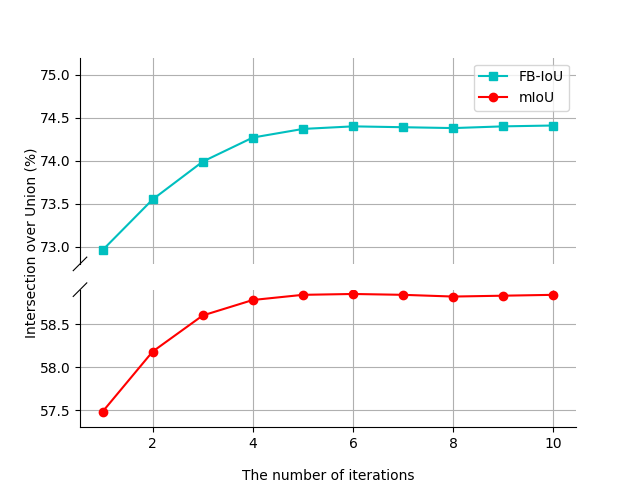}
\end{center}
\caption{Ablation study on evaluation iterations.}
\label{f1}
\end{figure}

To explore the effect of the number of iterations, we implement experiments of 1-shot setting with different iteration numbers on Pascal-$5^0$.
As shown in Figure \ref{f1}, both FB-IoU and mIoU increase monotonically with more iterations, and it takes around 5 iterations to obtain the converged result.

\section{Additional Qualitative Results}
\label{2}
\subsection{Visual Results on Pascal-\bm{$5^i$} and COCO-$20^i$}
\begin{figure*}[htbp]
\begin{center}
\includegraphics[width=17.3cm]{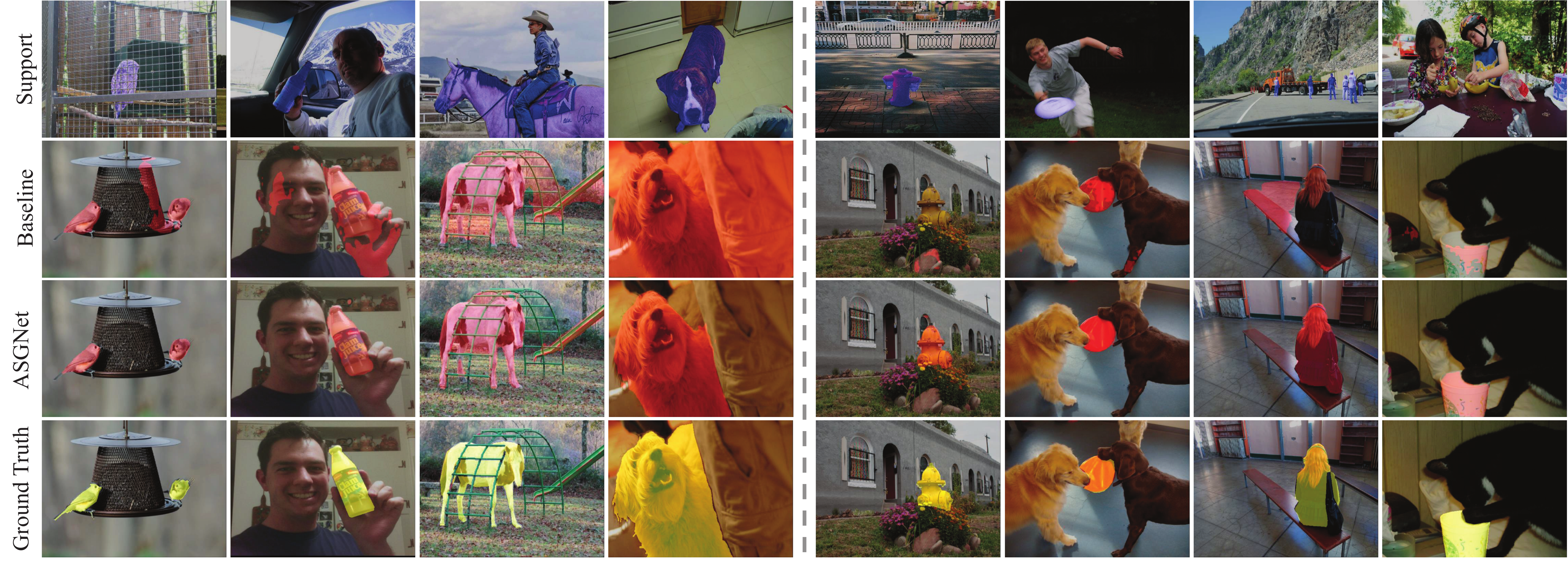}
\end{center}
\caption{Qualitative visualization of baseline (single prototype learning) and the proposed ASGNet. On the left are examples from Pascal-$5^i$ and the right ones are from COCO-$20^i$. Best viewed in color and zoom in.}
\label{f2}
\end{figure*}

In Figure \ref{f2}, we present more qualitative results in comparison to the single prototype baseline.
These qualitative results demonstrate that our model is capable of handling large variations in appearance, scale and shape between support and query images.
Compared with the baseline, we perform particularly better in occluded cases, e.g, column 3-6 of Figure \ref{f2}.

\begin{figure*}[htbp]
\begin{center}
\includegraphics[width=17.0cm]{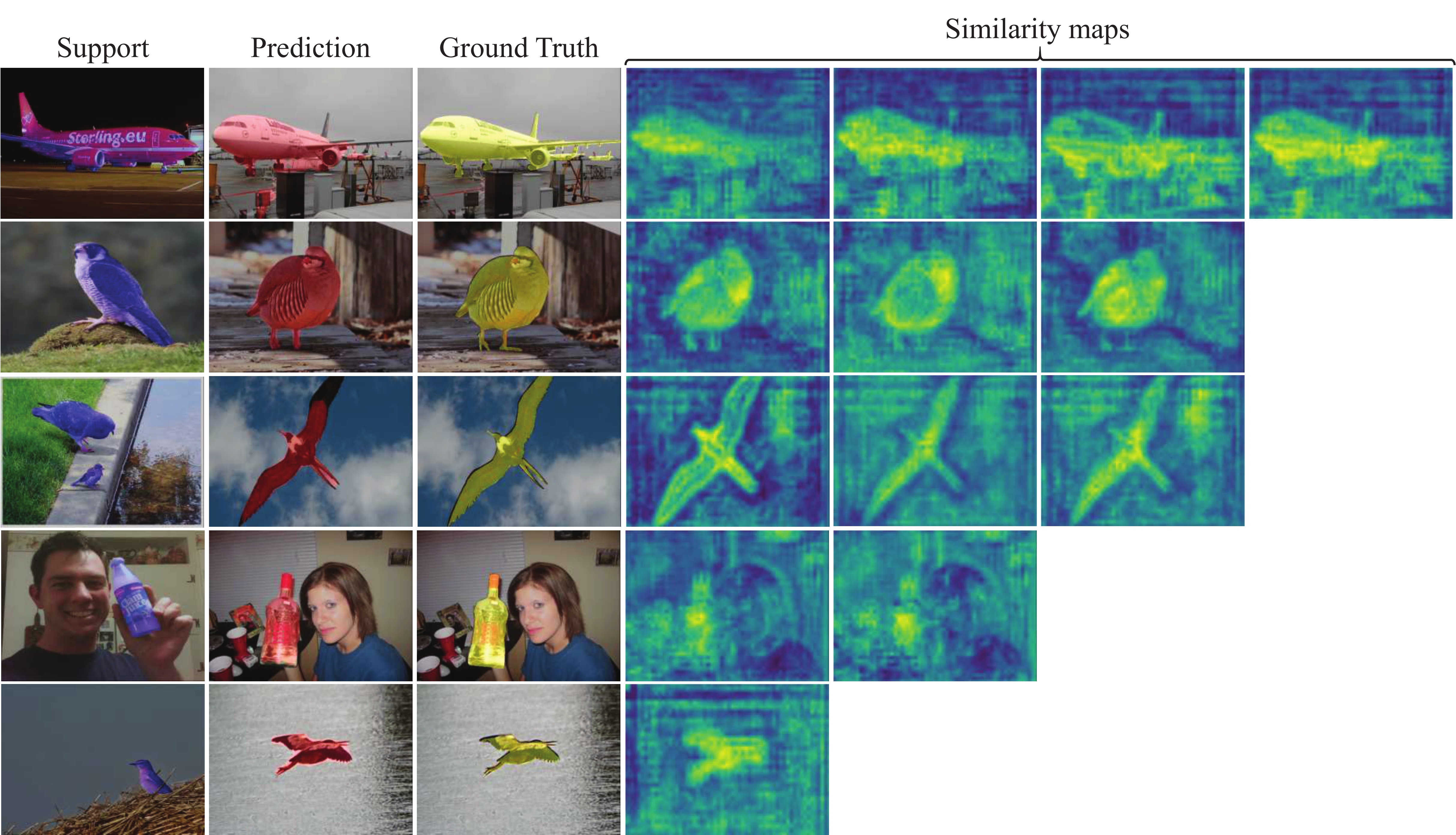}
\end{center}
\caption{Visualization of similarity maps on Pascal-$5^i$. The number of prototypes is determined by the size of support object. Best viewed in color and zoom in.}
\label{f3}
\end{figure*}

\subsection{Visualizations of Simiarity Map}
To better understand the proposed method, we visualize each similarity map, which is obtained by computing the cosine distance between each prototype and query feature.
As presented in Figure \ref{f3}, prototypes represent parts of the object with similar characteristics, which make the network more adaptive and discriminative. 

\end{document}